\documentclass[lettersize,journal]{IEEEtran}
\usepackage{amsmath,amsfonts}
\usepackage{algorithmic}
\usepackage{algorithm}
\usepackage{array}
\usepackage[caption=false,font=normalsize,labelfont=sf,textfont=sf]{subfig}
\usepackage{textcomp}
\usepackage{stfloats}
\usepackage{url}
\usepackage{verbatim}
\usepackage{graphicx}
\usepackage{cite}
\usepackage{caption2}
\usepackage{multirow}
\usepackage{booktabs}
\usepackage{makecell}
\def\BibTeX{{\rm B\kern-.05em{\sc i\kern-.025em b}\kern-.08em
    T\kern-.1667em\lower.7ex\hbox{E}\kern-.125emX}}
\usepackage{balance}
\begin{document}
\title{RTS-Mono: A Real-Time Self-Supervised Monocular Depth Estimation Method for Real-World Deployment}
\author{Zeyu Cheng, Tongfei Liu,~\IEEEmembership{Member,~IEEE}, Tao Lei,~\IEEEmembership{Senior Member,~IEEE}, 
	
	Xiang Hua, Yi Zhang and Chengkai Tang,~\IEEEmembership{Senior Member,~IEEE}
\thanks{This work was supported in part by National Natural Science Foundation of China under Grant 62271296, 62201334, 62171375; in part by The Innovation Capability Support Plan Project in Shaanxi Province under Grant 2025RS-CXTD-012; in part by Natural Science Basic Research Program of Shaanxi under Grant 2025JC-YBQN-800; in part by Scientific Research Program Funded by Shaanxi Provincial Education Department under Grant No.24JK0350; in part by Shenzhen Science and Technology Innovation Program under Grant JCYJ20220530161615033. (Corresponding author: Tao Lei.)}
\thanks{Zeyu Cheng, Tongfei Liu, and Tao Lei are with Shaanxi Joint Laboratory of Artificial Intelligence and the School of Electronic Information and Artificial Intelligence,  Shaanxi University of Science and Technology, Xi’an 710021, China (e-mail: chengzeyu@sust.edu.cn; liutongfei\_home@hotmail.com; leitao@sust.edu.cn). }
\thanks{Xiang Hua, Yi Zhang and Chengkai Tang are with the School of Electronics And Information, Northwestern Polytechnical University, 710129 Xi’an, China (e-mail: huaxiang1978@mail.nwpu.edu.cn; zhangyi@nwpu.edu.cn; cktang@nwpu.edu.cn).}
}

\markboth{arXiv preprint}%
{How to Use the IEEEtran \LaTeX \ Templates}

\maketitle

\begin{abstract}
Depth information is crucial for autonomous driving and intelligent robot navigation. The simplicity and flexibility of self-supervised monocular depth estimation are conducive to its role in these fields. However, most existing monocular depth estimation models consume many computing resources. Although some methods have reduced the model's size and improved computing efficiency, the performance deteriorates, seriously hindering the real-world deployment of self-supervised monocular depth estimation models in the real world. To address this problem, we proposed a real-time self-supervised monocular depth estimation method and implemented it in the real world. It is called RTS-Mono, which is a lightweight and efficient encoder-decoder architecture. The encoder is based on Lite-Encoder, and the decoder is designed with a multi-scale sparse fusion framework to minimize redundancy, ensure performance, and improve inference speed. RTS-Mono achieved state-of-the-art (SoTA) performance in high and low resolutions with extremely low parameter counts (3 M) in experiments based on the KITTI dataset. Compared with lightweight methods, RTS-Mono improved Abs Rel and Sq Rel by 5.6\% and 9.8\% at low resolution and improved Sq Rel and RMSE by 6.1\% and 1.9\% at high resolution. In real-world deployment experiments, RTS-Mono has extremely high accuracy and can perform real-time inference on Nvidia Jetson Orin at a speed of 49 FPS. Source code is available at https://github.com/ZYCheng777/RTS-Mono.
\end{abstract}


\section{Introduction}
\IEEEPARstart{D}{epth} information plays an important role in autonomous driving, intelligent robots, augmented reality, and virtual reality. Self-supervised monocular depth estimation does not rely on expensive depth labels and has excellent potential for real-world deployment in the above-mentioned fields. Real-world deployment in these fields requires the network model to have the ability to predict accurate depth information in real-time. However, the current self-supervised monocular depth estimation method cannot meet the requirements of real-time and high performance simultaneously, which seriously limits its real-world deployment in environments with limited computing resources. Therefore, how to make self-supervised monocular depth estimation predict depth in real time and accurately and deploy it in the real world is an urgent problem that needs to be solved.

Computational efficiency is a key metric for real-world deployment and directly affects the model's real-time performance. Generally speaking, the self-supervised monocular depth estimation network is an encoder-decoder architecture. This process includes a series of operations such as feature extraction, multi-scale downsampling, scale-wise upsampling, depth prediction, etc. These operations usually require many computing resources. In a real-world deployment, the model must process video stream information in real-time, and the processing speed needs to be 30 FPS or higher. Due to the complex structure of the network model and the large amount of calculation, it is challenging to meet the real-time requirements on devices with limited computing resources. This inference delay may cause the task to be unable to be completed on time, seriously affecting the safety and reliability of the real-world deployment of the model. Therefore, researchers are committed to designing a lighter and more efficient network architecture to reduce the amount of computation and parameters of the model and improve computing efficiency \cite{graham2021levit} \cite{wang2024binaryformer}.

Some progress has been made in the lightweight design of self-supervised monocular depth estimation. In order to reduce the network's operating resource requirements, the strategies of miniaturizing the network, removing redundant layers, and reducing the number of parameters are usually adopted in lightweight design. However, these measures limit the network's expression ability, decreasing depth estimation accuracy.

\begin{figure}[!t]
	\centering
	\includegraphics[width=3.5in]{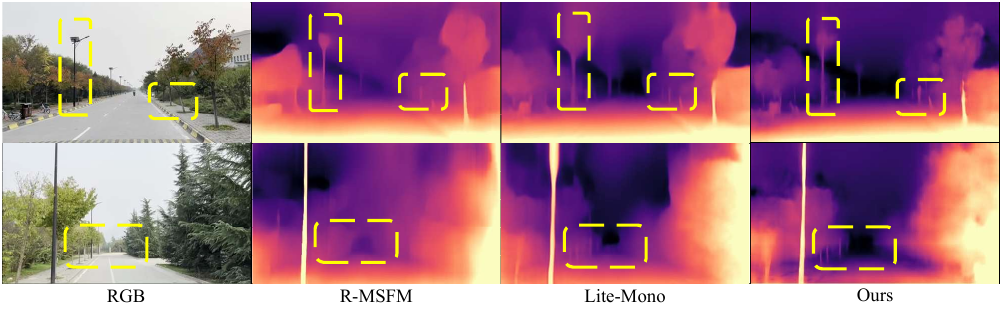}
	\caption{Comparison of the results of some lightweight self-supervised monocular depth estimation networks in real-world deployments.}
	\label{Ind}
\end{figure}

Lightweight networks are particularly inadequate when dealing with complex real-world scenes. For example, their estimation results often lack accuracy and consistency in areas with significant boundary details, texture changes, and dynamic light and shadow environments. This loss of accuracy affects the reconstruction quality and limits the network model's applicability in real-world deployment. As shown in Fig. \ref{Ind}, the current lightweight monocular depth estimation network has varying degrees of attenuation in terms of boundary details and scene depth accuracy in real-world deployment.

The depth estimation task needs to capture rich features to model the geometric and semantic information of the scene accurately. However, due to the reduction in the number of parameters and computing power, the lightweight network cannot learn sufficiently detailed feature expressions. In order to improve the network's expression ability, the number of network parameters and the amount of computing power need to be increased. As seen from TABLE \ref{tab:Ind}, the reduction in network size and the increase in computing efficiency will sacrifice some performance, and the performance improvement will sacrifice computing efficiency. Therefore, finding a better balance between computing efficiency and network performance still requires more research.

\begin{table}[!t]
	\caption{Comparison of the proportion of decoder parameters to the full model parameters.}
	\label{tab:Ind}
	\centering
	\setlength{\tabcolsep}{5.5mm}{
		\begin{tabular}{c c c c }
			\specialrule{0em}{3pt}{3pt}
			\toprule[1pt]
			
			Method  &  \#Params & FPS& RMSE \\			
			\specialrule{0.05em}{1pt}{5pt}
			RA-Depth \cite{he2022ra}    &10.0 M &12&4.216 \\
			MonoViT-tiny \cite{zhao2022monovit}     &10.3 M &14&4.459   \\
			R-MSFM \cite{zhou2021r}    &3.8 M &21&4.704    \\
			Lite-Mono \cite{zhang2023lite}    &3.1 M &45&4.561   \\
			GuideDepth \cite{rudolph2022lightweight}   &5.8 M &51&5.194    \\
			RT-MonoDepth \cite{feng2024real}   &2.8 M&83&4.985    \\
			RTIA-Mono \cite{zhao2025rtia}   &1.4 M&49&4.810    \\
			
			\bottomrule[1pt]
	\end{tabular}}
\end{table}

In this paper, we propose a real-time self-supervised monocular depth estimation method, RTS-Mono. It consists of a lightweight and efficient encoder and decoder to ensure depth estimation accuracy at the lowest cost. We deploy it on intelligent UAV systems for real-world, real-time depth estimation. A large number of dataset experiments and real-world deployment experiments show that RTS-Mono is highly competitive compared to lightweight methods.

Specifically, our contributions are as follows:
\begin{enumerate}
	\item {We designed a lightweight and efficient self-supervised monocular depth estimation encoder-decoder architecture, with the encoder Lite-Encoder and the decoder a multi-scale sparse fusion framework. This architecture maximizes computational efficiency and guarantees performance by removing redundant fusion processes.}
	\item {A large number of experiments were conducted on the challenging dataset KITTI. The results show that our method achieves SoTA performance at low and high resolutions with extremely low parameters. Compared with the current lightweight SoTA method, it can improve the performance of Abs Rel, Sq Rel, and RMSE metrics by 5.6\%, 9.8\%, and 4.8\% at low resolution and 6.1\% and 1.9\% at high resolution, respectively.}
	\item {A large number of experiments were conducted in real urban traffic environments. We built an intelligent UAV depth perception system and deployed multiple methods including RTS-Mono on it, and conducted real-time depth estimation experiments in the real world. The results show that our method can estimate depth in real scenes in real time, with an inference speed of up to 49 FPS, and surpasses all lightweight self-supervised monocular depth estimation methods in performance, making it highly competitive.}
\end{enumerate}

\section{Related work}
\subsection{Lightweight deep learning network architecture}
Convolutional neural networks (CNNs) focus on local areas of input data through local receptive fields, extract low-level features through convolutional layers, and then stack convolutional layers layer by layer to extract higher-level semantic information. This hierarchical feature extraction method enables CNNs to handle input data with spatial structure, such as images and videos. This has made CNNs the mainstream framework for computer vision for quite a long time. However, CNNs cannot capture long-range dependencies, especially in scenarios where global information needs to be modeled, and their performance is limited. Transformers can directly capture the global dependencies of input data, which makes them particularly suitable for processing tasks that require global contextual information, such as natural language processing (NLP) and complex visual tasks, and well compensate for the limitations of CNNs in global modeling. Compared with CNNs, Transformers usually have larger parameters and higher expressive power, but they also require higher computing resources, which hinders their deployment on embedded devices. Therefore, lightweighting of high-performance networks has become an urgent problem to be solved. The MobileNet \cite{howard2017mobilenets} series of networks based on CNNs are typical representatives of lightweight models. They proposed depth-wise separable convolution, which decomposes the traditional convolution operation into two steps: depth-wise convolution and point-wise convolution, thereby reducing the number of parameters and computational costs. Subsequent versions of MobileNet \cite{sandler2018mobilenetv2} \cite{howard2019searching} introduced modules such as Inverted Residual Block, Linear Bottleneck, adaptive convolution, and attention mechanism to further improve performance and computational efficiency. On the Transformer side, DeiT \cite{touvron2021training} (Data-efficient Image Transformer) optimizes the training process by introducing distilled tokens, reducing data requirements while maintaining high performance. Swin-Transformer \cite{liu2021swin} uses a sliding window mechanism for self-attention calculations, significantly reducing computational complexity while retaining global feature modeling capabilities, making it suitable for high-resolution tasks. LeViT \cite{graham2021levit} achieves fast reasoning by combining deep convolution and Transformer modules, making it a low-latency visual Transformer. CvT \cite{wu2021cvt} (Convolutional Vision Transformer) combines convolution operations with Transformer self-attention, retaining the local feature extraction capabilities of convolution and introducing the global modeling capabilities of Transformers, combining efficiency and performance. TinyViT \cite{wu2022tinyvit} is specifically designed for edge devices and resource-constrained scenarios, significantly reducing computational complexity and storage requirements while ensuring performance.

These lightweight network architectures reduce the complexity of the model through novel design and optimization strategies. They are suitable for resource-constrained devices and application scenarios with high real-time requirements. However, there are still some problems in applying these lightweight network frameworks to self-supervised monocular depth estimation. In order to reduce the amount of computation and the number of parameters, lightweight networks usually reduce the number of layers, width, and feature channels of the model. This simplification may lead to insufficient feature extraction capabilities of the monocular depth estimation network, especially in complex scenes in the real world, where it is challenging to capture sufficient local and global information required for depth estimation. Moreover, the depth estimation task needs to capture information at different scales to understand the geometric structure of the scene, but the multi-scale feature processing capability of lightweight networks is weak, leading to uneven depth estimation of distant and near targets. This series of problems affects the deployment of monocular depth estimation tasks in the real world.
\subsection{Self-supervised monocular depth estimation}
Using cameras to obtain depth information is an important way for intelligent agents such as autonomous driving systems, robots, and UAVs to perceive and understand the external environment. The acquisition of depth information relies on LiDAR, stereo cameras, and multi-sensor fusion solutions, which are expensive and complex. Using monocular cameras for depth estimation provides a low-cost, lightweight solution, which is particularly suitable for commercial autonomous vehicles and small UAVs. Monocular cameras can significantly reduce costs and facilitate the promotion and popularization of products. Although monocular depth estimation is an ill-posed problem, the development of deep learning makes it possible to solve this problem with data-driven methods. In addition, self-supervised methods for monocular depth estimation can be based on data-driven without relying on expensive depth labels as supervisory signals, which is more conducive to real-world deployment. Therefore, the real-world deployment of self-supervised monocular depth estimation on intelligent agents has a strong potential for application.
\begin{figure*}[!t]
	\centering
	\includegraphics[width=7in]{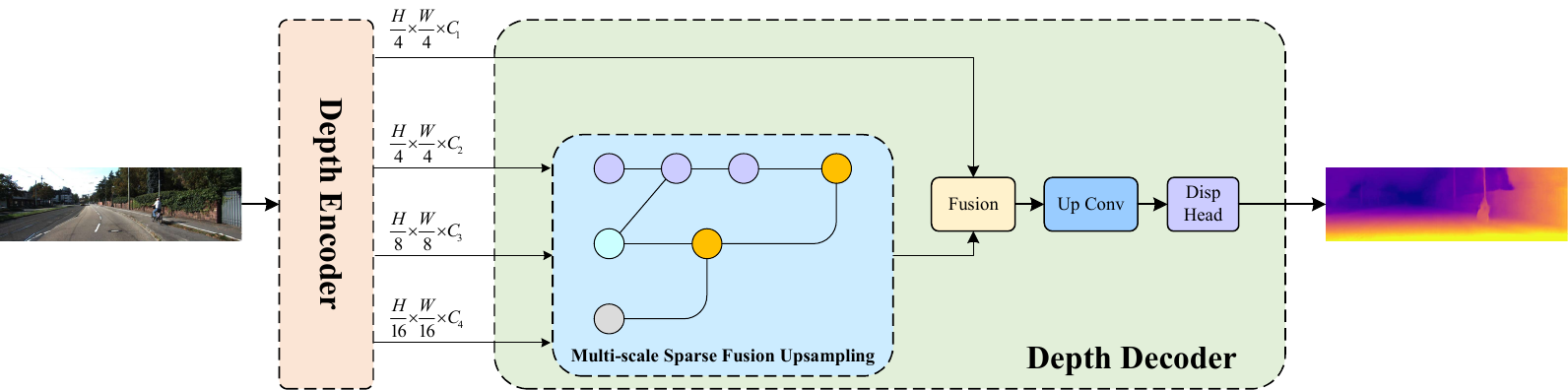}
	\caption{Network architecture overview.}
	\label{overall}
\end{figure*}

Self-supervised monocular depth estimation effectively solves the problem that the depth estimation network depends on depth labels, and outstanding works have been proposed one after another \cite{godard2019digging} \cite{bian2019unsupervised} \cite{klingner2020self} \cite{johnston2020self} \cite{yan2021channel}. However, these works did not consider the real-world deployment of intelligent agents, so their performance and computational efficiency are not balanced. MonoViT \cite{zhao2022monovit} compensates for the shortcomings of the two architectures in the monocular depth estimation task by fusing ordinary convolution with the Transformer architecture while maintaining computational efficiency. Lyu proposed Lite-HR-Depth \cite{lyu2021hr} with MobileNetV3 \cite{howard2019searching} as the encoder while proposing HR-Depth, reducing the number of parameters and computing resources of the model. RA-Depth \cite{he2022ra} ensures the consistency of depth between different scales by introducing multi-scale depth consistency loss to the model and proposes a dual high-resolution network (Dual HRNet) to improve the accuracy of depth prediction at a lower cost. R-MSFM \cite{zhou2021r} uses a part of ResNet18 \cite{he2016deep} as an encoder to ensure the lightness of the architecture and proposes a multi-scale feature modulation module to ensure performance. GuideDepth \cite{rudolph2022lightweight} proposes a Guided Upsampling Block in the decoder, which uses input images of different resolutions to guide the decoder in upsampling the feature representation and predicting the depth. It has good real-time performance. RT-MonoDepth \cite{feng2024real} proposes a lightweight pyramid encoder and a simplified depth decoder to maximize the network's inference speed. Lite-Mono \cite{zhang2023lite} designs an efficient encoder and forms an encoding-decoding architecture with the decoder of MonoDepth2 \cite{godard2019digging}. The architecture has a Consecutive Dilated Convolutions module and a Local-Global Features Interaction module to improve performance and ensure computational efficiency. RTIA-Mono \cite{zhao2025rtia} improves upon \cite{zhang2023lite} by designing the Global Local Information Aggregation module and the Directional Feature Enhancement module, further reducing the network size and improving computational efficiency. However, although some of these lightweight methods reduce computational complexity and achieve real-time performance, they sharply deteriorate depth prediction accuracy. Some methods have a degree of accuracy guarantee but cannot achieve real-time prediction. Therefore, these methods cannot balance performance and computational efficiency, which is not conducive to the real-world deployment of network models.

We note these problems in the real-world deployment of lightweight self-supervised monocular depth estimation. In this paper, we propose a high-performance and efficient decoder with a multi-scale sparse fusion architecture, combined with a lightweight encoder, to form a high-performance encoder-decoder architecture capable of real-time inference. This architecture surpasses existing lightweight networks and balances network performance and computational efficiency.

\section{Method}
\subsection{Motivation}
The ultimate goal of monocular depth estimation is to be freely deployed on intelligent agents such as vehicles, robots, and UAVs and to assist the agent in perceiving and navigating the environment. This requires the model to have accurate depth prediction capabilities and real-time computational efficiency. Although the current monocular depth estimation model has achieved lightweight, \cite{zhou2021r} only focuses on the lightweight encoder; that is, it uses a partially simplified ResNet18 but does not consider the computational efficiency of the decoder, which results in a small model size, but the inference speed is still slow; \cite{rudolph2022lightweight} \cite{feng2024real} has achieved lightweight network overall and has good real-time performance, but there is a significant gap in the accuracy of depth prediction compared with general methods; \cite{zhang2023lite} has achieved lightweight in the overall architecture of the network, but its decoder lacks sufficient local and global modeling, so the accuracy of the network model cannot be guaranteed. Therefore, these methods cannot balance network performance and computational efficiency and are not conducive to the real-world deployment of the model. Based on this problem, we focus on designing a high-performance real-time encoder-decoder architecture that can be deployed, combining PoseNet \cite{godard2019digging} and existing loss functions for self-supervised training to predict the depth information of continuous image sequences in real-time with extremely low computational overhead, and strive to achieve a balance between performance and computational efficiency.
\subsection{Depth Network}
The entire network architecture is simple to achieve and extremely lightweight. The overall architecture is shown in Fig \ref{overall}. The network consists of two parts: encoder and decoder. The encoder uses Lite-Encoder  \cite{zhang2023lite}, a lightweight hierarchical multi-stage network that uses a hybrid architecture of CNNs and Transformers, effectively reducing the number of parameters and calculations. Specifically, the input RGB image $\textbf{\textit{I}}\in {{\mathbb{R}}^{H\times W\times 3}}$ passes through each stage of the encoder to obtain four features of different resolutions from shallow to deep layers:
\begin{equation}\label{eq1}
	\begin{aligned}
		& {{\textbf{\textit{F}}}_{0}}\in {{\mathbb{R}}^{\frac{H}{4}\times \frac{W}{4}\times {{C}_{1}}}} \\ 
		& {{\textbf{\textit{F}}}_{1}}\in {{\mathbb{R}}^{\frac{H}{4}\times \frac{W}{4}\times {{C}_{2}}}} \\ 
		& {{\textbf{\textit{F}}}_{2}}\in {{\mathbb{R}}^{\frac{H}{8}\times \frac{W}{8}\times {{C}_{3}}}} \\ 
		& {{\textbf{\textit{F}}}_{3}}\in {{\mathbb{R}}^{\frac{H}{16}\times \frac{W}{16}\times {{C}_{4}}}} \\ 		 		
	\end{aligned}
\end{equation}
where ${{C}_{1}}$, ${{C}_{2}}$, ${{C}_{3}}$ and ${{C}_{4}}$ represent the number of channels of the feature, and we set them to 48, 48, 80, and 128 respectively. Subsequently, these features of different resolutions enter the decoder.

\begin{figure}[!t]
	\centering
	\includegraphics[width=3.5in]{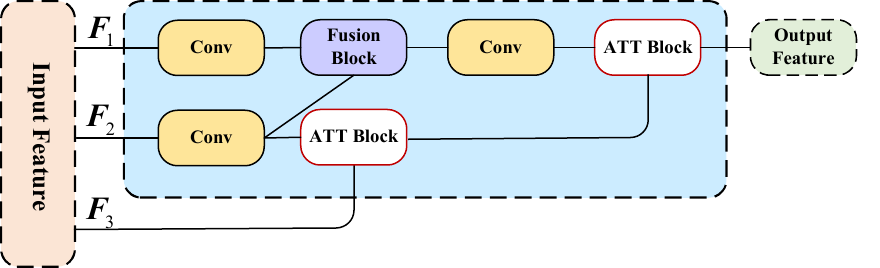}
	\caption{Multi-scale sparse fusion upsampling module.}
	\label{SparseFusion}
\end{figure}
\begin{figure}[!t]
	\centering
	\includegraphics[width=3.5in]{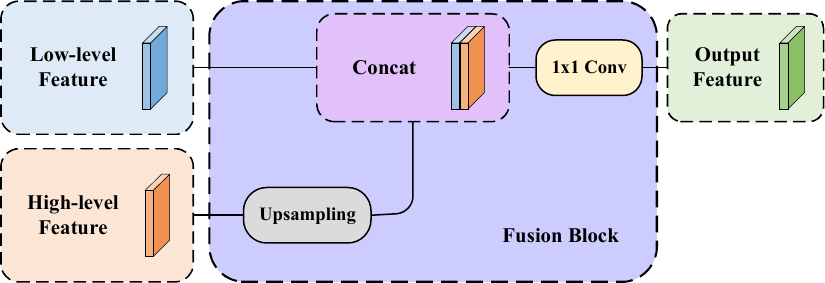}
	\caption{Fusion block.}
	\label{FusionModule}
\end{figure}

The decoder also adopts a lightweight design concept. Starting from the features of each scale, the features are divided into low-level shallow features and high-level abstract features. Only the high-level abstract features are sent to the layer-wise upsampling module to ensure sufficiently low computational complexity and few parameters. The low-level shallow features are directly integrated with the output features of the layer-wise upsampling module to ensure feature integrity and performance.

Specifically, ${{\textbf{\textit{F}}}_{0}}$, ${{\textbf{\textit{F}}}_{1}}$, ${{\textbf{\textit{F}}}_{2}}$, and ${{\textbf{\textit{F}}}_{3}}$ obtained from the encoding end are divided into low-level shallow features ${{\textbf{\textit{F}}}_{0}}$ and high-level abstract features ${{\textbf{\textit{F}}}_{1}}$, ${{\textbf{\textit{F}}}_{2}}$, and ${{\textbf{\textit{F}}}_{3}}$. The high-level abstract features enter the multi-scale sparse fusion upsampling module, as shown in Fig. \ref{SparseFusion}. This module also adopts a sparse fusion and lightweight design, containing only three convolution operations, one fusion block, and two attention blocks. First, ${{\textbf{\textit{F}}}_{1}}$ and ${{\textbf{\textit{F}}}_{2}}$ are subjected to a convolution operation with a convolution kernel size $3\times 3$ (represented as Conv in Fig. \ref{SparseFusion}) to obtain ${{\textbf{\textit{F}}}_{1}}^{\prime }$ and ${{\textbf{\textit{F}}}_{2}}^{\prime }$ , then enter two fusion upsampling modules. ${{\textbf{\textit{F}}}_{2}}^{\prime }$ and ${{\textbf{\textit{F}}}_{3}}$ enter the ATT Block for fusion upsampling to obtain ${{\textbf{\textit{F}}}_{2}}^{\prime \prime }$, and ${{\textbf{\textit{F}}}_{1}}^{\prime }$ and ${{\textbf{\textit{F}}}_{2}}^{\prime }$ enter the Fusion Block for fusion upsampling to obtain ${{\textbf{\textit{F}}}_{1}}^{\prime \prime }$. ATT Block is a fusion method of fusion Squeezeand-Excitation (fSE) \cite{lyu2021hr}. The Fusion Block is shown in Fig. \ref{FusionModule}. We refer to the structure of fSE for the design of the Fusion Block. Two different levels of features enter the fusion module. The high-level features are upsampled, and the low-level features reach a resolution, and then the two features are merged. In fSE, after the two features are merged, a series of attention mechanism operations are performed. In Fusion Block, we use point-wise convolution ($1\times 1$ Conv in Fig. \ref{FusionModule}) instead of this process to reduce the amount of calculation and parameters. It should be noted here that the high-level features and low-level features in Fig. \ref{FusionModule} are only relative concepts of the two features, not the low-level shallow features and high-level abstract features mentioned above. Bringing this module into the inference process, we get
\begin{equation}\label{eq2}
	{{{\textbf{\textit{F}}}''}_{1}}=Con{{v}_{1\times 1}}\left( Concat\left( {{{{\textbf{\textit{F}}}'}}_{1}},upsampling\left( {{{{\textbf{\textit{F}}}'}}_{2}} \right) \right) \right).
\end{equation}

The deepest feature, ${{\textbf{\textit{F}}}_{3}}$, uses the more expressive ATT Block in the fusion and upsampling process. We use this architecture to improve the network's global understanding and overall control of the scene, thereby ensuring the performance of the network architecture. After ${{\textbf{\textit{F}}}_{1}}$ is fused with ${{\textbf{\textit{F}}}_{2}}$ through the Fusion Block, a $3\times 3$ convolution operation is performed, and then ${{\textbf{\textit{F}}}_{2}}^{\prime \prime }$ is fused through the ATT Block to obtain the output feature $\textbf{\textit{F}}$. The calculation process is as follows:
\begin{equation}\label{eq3}
	\textbf{\textit{F}}=ATT\left( Con{{v}_{3\times 3}}\left( {{{{\textbf{\textit{F}}}''}}_{1}} \right),{{{{\textbf{\textit{F}}}''}}_{2}} \right).
\end{equation}

The output feature $\textbf{\textit{F}}$ is then fused with the low-level shallow feature ${{\textbf{\textit{F}}}_{0}}$ and a series of operations to obtain the final output. In Fig. \ref{overall}, Upconv is a convolution operation with a convolution kernel of $3\times 3$, which is used to reduce the dimension of the newly fused features. Disp Head contains a convolution operation, an upsampling and a Sigmoid. The calculation process is as follows:
\begin{equation}\label{eq4}
	\resizebox{1\hsize}{!}{$
		{{\textbf{\textit{F}}}_{out}}=Sigmoid\left( upsampling\left( Con{{v}_{3\times 3}}\left( Concat\left( {{\textbf{\textit{F}}}_{0}},upsampling\left( \textbf{\textit{F}} \right) \right) \right) \right) \right).
		$}
\end{equation}

\subsection{Discussion of Multi-scale Sparse Fusion Decoder} \label{sec:discuss}
\begin{figure*}[!t]
	\centering
	\includegraphics[width=6.8in]{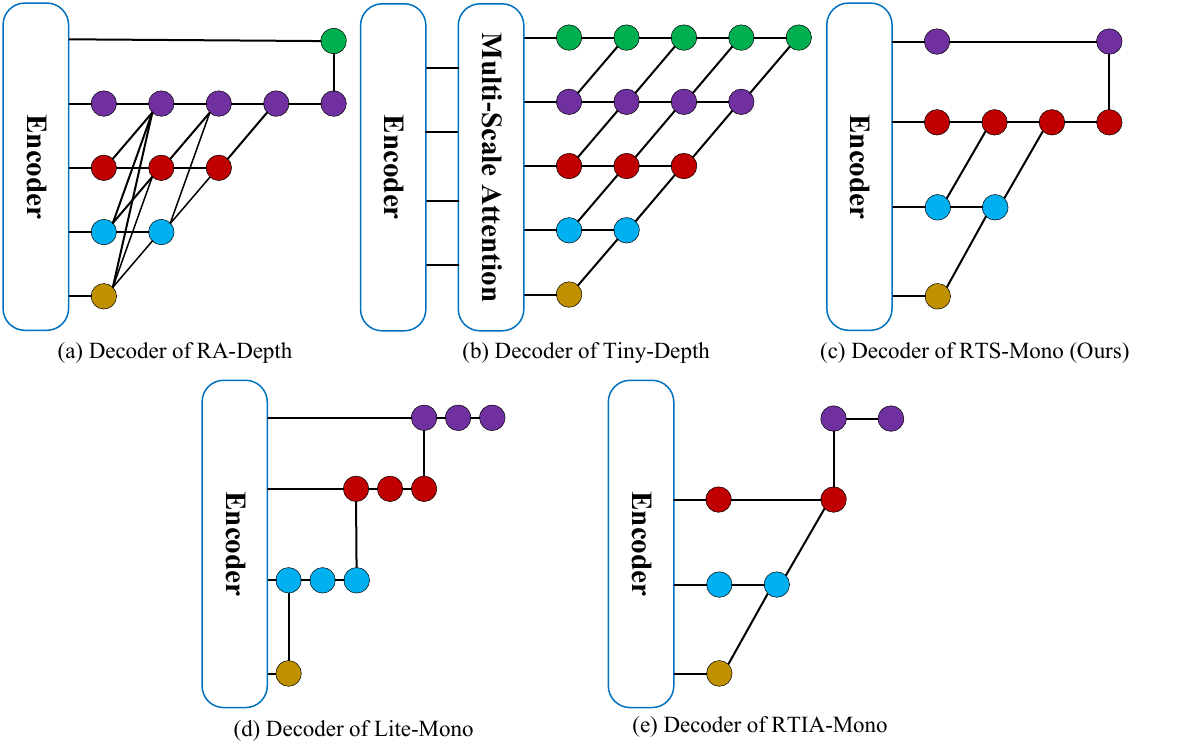}
	\caption{Decoder structure comparison.}
	\label{decoder}
\end{figure*}
After introducing the encoder-decoder architecture of our deep prediction network, we discuss the decoder structure related to our decoder in this section to illustrate our design intention further. Fig. \ref{decoder} is the abstract framework of the decoder. The circular blocks in the figure represent a series of operations such as convolution, fusion, and sampling. Fig. \ref{decoder}(a) shows the decoder of the quasi-lightweight network RA-Depth \cite{he2022ra}. It can be seen that five features enter the decoder. The following four features are integrated with all deeper features at each step in the upsampling process. After the sampling fusion is completed, they are merged with the shallowest features. This design is of great help to the inference performance. However, each sampling must integrate all high-level features. This design makes the network structure complex, ultimately affecting the network's inference speed. A sparse fusion design is adopted in the TinyDepth \cite{cheng2024tinydepth} decoder (Fig. \ref{decoder}(b)); each sampling process reduces the number of fused features and only fuses with adjacent deep features, reducing the complexity of the network structure. In order to ensure network performance, the encoder outputs four features for multi-scale attention processing and five different levels of features for sparse fusion upsampling. Such scale-wise attention and multiple fusion operations of shallow features slow the inference speed. The decoder of our method draws on the advantages of RA-Depth and TinyDepth and follows the lightweight design requirements. As shown in Fig. \ref{decoder}(c), the encoder outputs four features that directly enter the decoder, and the three features with deeper layers enter the sparse fusion sampling stage. The results of the fusion sampling are directly fused with the shallow features. This series of operations can minimize the computational complexity, improve the inference speed, and ensure performance. LiteMono \cite{zhang2023lite}, also a lightweight network, reduces the input features of the decoder based on the classic work Monodepth2 to reduce the computational overhead of the decoder (Fig. \ref{decoder})(d)). Monodepth2 reduces the interaction between high-level and low-level features at the decoding end to ensure the inference speed, resulting in the inability to utilize global context information for depth prediction fully. On this basis, LiteMono further reduces the input features to improve computational efficiency, sacrificing network performance. As for RTIA-Mono \cite{zhao2025rtia}, although its architecture is similar to our method, it has a certain degree of simplification in the key deep-shallow feature fusion and upsampling processes. These result in it having a lower number of parameters and computational complexity than our method, but the network's expressive power is far behind our method.

In summary, our method starts from the network architecture, maximizes the inference speed, and guarantees the performance through a series of operations such as feature partitioning and sparse fusion. The specific comparative analysis of the methods mentioned in this section will be carried out in the experiments section.

\subsection{Self-supervised Learning}
In order to improve the generalization ability of the lightweight network model, our network is trained using a self-supervised method, so it is necessary to generate a self-supervised signal for iterative training. We use the relationship between consecutive frame sequences of monocular images to generate self-supervised signals without relying on depth labels and convert the monocular depth estimation problem into an image reconstruction problem. Specifically, $\textbf{\textit{I}}$ and $\hat{\textbf{\textit{I}}}$ are two temporally adjacent frames. We take  $\textbf{\textit{I}}$ as the target frame and $\hat{\textbf{\textit{I}}}$ as the source frame. Meanwhile, $\textbf{\textit{D}}$ is the depth of the target frame, $\textbf{\textit{T}}$ is the camera pose between the target frame and the source frame, and the reconstructed target frame can be expressed as
\begin{equation}\label{eq5}
	\tilde{\textbf{\textit{I}}}=\hat{\textbf{\textit{I}}}\left\langle \text{proj}\left( \textbf{\textit{D}},\textbf{\textit{T}},\textbf{\textit{K}} \right) \right\rangle,
\end{equation}
where $\textbf{\textit{K}}$  is the camera internal parameter, $\text{proj(}\bullet \text{)}$ and $\left\langle \bullet  \right\rangle $ represent coordinate projection and sampling operations respectively, thus constructing the image reconstruction loss
\begin{equation}\label{eq6}
	{{L}_{p}}=\underset{i}{\mathop{\min }}\,\left\{ \alpha \cdot \frac{1-\text{SSIM}\left( \textbf{\textit{I}},{{{\tilde{\textbf{\textit{I}}}}}_{i}} \right)}{2}+\left( 1-\alpha  \right){{\left\| \textbf{\textit{I}}-{{{\tilde{\textbf{\textit{I}}}}}_{i}} \right\|}_{1}} \right\},
\end{equation}
where $i\in \left[ -1,1 \right]$, ${{\tilde{\textbf{\textit{I}}}}_{i}}$ is a synthetic image, which is reconstructed from the target frame $\textbf{\textit{I}}$ and the previous and next frames adjacent in time. The image reconstruction loss also includes SSIM \cite{wang2004image} and ${{L}_{1}}$ \cite{godard2019digging} , and the image reconstruction loss takes the minimum value of the photometric error function $phe\left( \bullet  \right)$. 

In addition, we deploy edge-aware smoothness loss \cite{sun2024dynamo} to ensure the continuity of the estimated disparity. The calculation process is
\begin{equation}\label{eq7}
	{{L}_{s}}=\left| {{\partial }_{x}}{{\textbf{\textit{D}}}^{*}} \right|{{e}^{{{\partial }_{x}}\textbf{\textit{I}}}}+\left| {{\partial }_{y}}{{\textbf{\textit{D}}}^{*}} \right|{{e}^{{{\partial }_{y}}\textbf{\textit{I}}}},
\end{equation}
where ${{\textbf{\textit{D}}}^{*}}=\textbf{\textit{D}}/\bar{\textbf{\textit{D}}}$ represents the average normalized inverse depth and ${{\partial }_{x}}$ and ${{\partial }_{y}}$ represent the gradients in the horizontal and vertical directions, respectively.

We also deploy a cross-scale depth consistency loss ${{L}_{d}}$ \cite{he2022ra} to ensure depth consistency between different scales. Its core idea is to scale image $\textbf{\textit{I}}$ into input images ${{\textbf{\textit{I}}}_{L}}$, ${{\textbf{\textit{I}}}_{M}}$ and ${{\textbf{\textit{I}}}_{H}}$ of different scales and then enter the network to obtain depth maps ${{\textbf{\textit{D}}}_{L}}$, ${{\textbf{\textit{D}}}_{M}}$, ${{\textbf{\textit{D}}}_{H}}$ of different scales and use the common areas of these depth maps to calculate the loss. The process is as follows:
\begin{equation}\label{eq8}
	\begin{split}
		{{L}_{d}} &=L_{D}^{LM}+L_{D}^{MH} \\
		&=\left( 1-\alpha  \right)\left( {{\left\| {{{\tilde{\textbf{\textit{D}}}}}_{L}}-{{{\tilde{\textbf{\textit{D}}}}}_{M}} \right\|}_{1}}+{{\left\| {{{\hat{\textbf{\textit{D}}}}}_{M}}-{{{\hat{\textbf{\textit{D}}}}}_{H}} \right\|}_{1}} \right) \\
		&+\frac{\alpha }{2}\left( SSIM\left( {{{\tilde{\textbf{\textit{D}}}}}_{L}},{{{\tilde{\textbf{\textit{D}}}}}_{M}} \right)+SSIM\left( {{{\hat{\textbf{\textit{D}}}}}_{M}},{{{\hat{\textbf{\textit{D}}}}}_{H}} \right) \right)+\alpha ,
	\end{split}
\end{equation}
where ${{\tilde{\textbf{\textit{D}}}}_{M}}$ and ${{\tilde{\textbf{\textit{D}}}}_{L}}$  are the common area of ${{\textbf{\textit{D}}}_{M}}$, and the depth consistency losses $L_{D}^{LM}$ and $L_{D}^{MH}$  at different scales are obtained. Then, summed up, the cross-scale depth consistency loss ${{L}_{d}}$ is obtained. Therefore, our loss function is
\begin{equation}\label{eq9}
	Loss=\gamma \cdot {{L}_{p}}+\beta \cdot {{L}_{s}}+\lambda \cdot {{L}_{d}},
\end{equation}
where ${{L}_{s}}$ is the edge-aware smoothness loss, ${{L}_{d}}$ is the cross-scale depth consistency loss, and $\gamma $, $\beta $, $\lambda $ are hyperparameters.

\begin{figure*}[!t]
	\centering
	\includegraphics[width=4.8in]{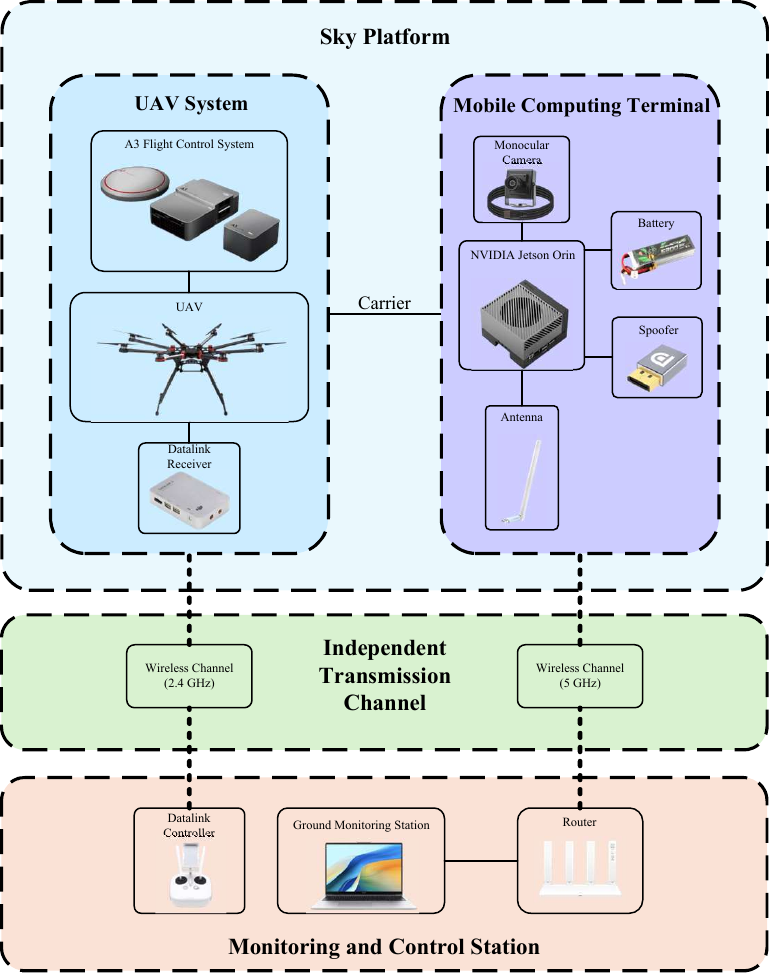}
	\caption{System hardware framework.}
	\label{System_details}
\end{figure*}

\section{Intelligent UAV depth perception system}\label{sec4}
In this section, we propose an intelligent UAV depth perception system based on our method RTS-Mono. The system deploys RTS-Mono on the mobile computing terminal NVIDIA Jetson Orin and uses UAV as a carrier to achieve real-time depth perception of real-world urban environments.

We have designed a comprehensive hardware solution from the sky end and transmission end to the ground end to ensure the efficient, stable, and low-latency operation of the entire intelligent UAV depth perception system. The hardware system architecture is shown in the Fig.  \ref{System_details}.  In  Fig.  \ref{System_details}, the main body of the sky platform is an octocopter UAV equipped with a DJI A3 flight control system. The mobile computing terminal consists of NVIDIA Jetson Orin, a monocular camera, a battery, a graphics card spoofer, and an antenna. Orin is independently powered by a high-performance battery. It is not only small in size and light in weight, making it easy to integrate into the UAV system, but it also does not rely on the UAV battery, which simplifies the power connection process and improves the overall safety of the system. The battery can give Orin with a maximum power supply of 12V/5A, ensuring the mobile computing terminal can work stably under high loads. In terms of sensors, a high-definition camera is selected as a monocular vision sensor for real-time image acquisition. A USB interface is used to prevent the interface from interfering with the frequency band of the UAV GPS antenna. In addition, the graphics card spoofer is used to simulate the display data to reduce the data delay between the ground station and the sky platform. The antenna enhances the signal and has two working frequency bands: 2.4 GHz and 5 GHz.

The middle part of Fig.  \ref{System_details} is the signal transmission stage. This stage adopts an independent transmission design and sets two transmission channels of different frequency bands, 2.4 GHz and 5 GHz. The Datalink controller connects to the Datalink receiver of the UAV system through the 2.4 GHz channel to control the UAV. The ground monitoring station connects to the mobile computing terminal through the 5 GHz channel for real-time monitoring. The design of independent transmission channels minimizes interference between channels and ensures the stable operation of the entire system.

The bottom of Fig.  \ref{System_details} is the ground monitoring and control station. On the one hand, the Datalink controller controls the flight of the UAV through an independent transmission link. On the other hand, the ground monitoring station is connected to the mobile computing terminal through a router and performs real-time control and monitoring through the remote control software NoMachine\cite{fenton2021virtual}. The depth predicted by the mobile computing terminal is presented to the ground monitoring station through the independent transmission channel for the first time.

\section{Experiments}
In this section, we conducted many experiments on the KITTI dataset and provided detailed comparison results to verify the theoretical performance of our method. We deployed our method on the intelligent UAV system we built and conducted experiments in the real world to verify the deployment performance of our method.  Finally, the ablation study verified the effectiveness of our method.
\subsection{Implementation details}
In terms of datasets, we use Eigen's split to train and test the KITTI dataset \cite{geiger2013vision}, which contains 39,810 training images and 697 test images. We limit the maximum depth to 80 m. In terms of the network model, we divide RTS-Mono into two versions, XS and S, according to the encoder specification and read the encoder's pre-trained weights on ImageNet. We set the value of $\alpha $ in Equation \ref{eq6} and \ref{eq8} to 0.85 and the values of $\gamma $,  $\beta $, and $\lambda $ in Equation \ref{eq9} to 1.0, 0.001, and 1.0, respectively. Our network model was experimented with on NVIDIA RTX A6000. The Adam optimizer \cite{kingma2014adam} is used during the training process, the initial learning rate is set to ${{10}^{-4}}$ , and the cosine learning rate \cite{loshchilov2016sgdr} is deployed. The batch size is set to 12, and the epochs are set to 95. When training the KITTI dataset, the image size is adjusted to $640\times192$ and $1024\times320$. For performance evaluation, we use widely adopted metrics \cite{eigen2014depth}, namely absolute relative error (AbsRel), square relative error (Sq Rel), root mean square error (RMSE), root mean square logarithmic error (RMSE log), and accuracy (${{\delta }_{1}}<1.25$, ${{\delta }_{2}}<{{1.25}^{2}}$, ${{\delta }_{3}}<{{1.25}^{3}}$). These metrics are calculated as follows:
\begin{equation}\label{eq11}
	\begin{aligned}
		& \text{RMSE}=\sqrt{\frac{1}{\left| N \right|}{{\sum\limits_{i\in N}{\left\| {{d}_{i}}-{{{\tilde{d}}}_{i}} \right\|}}^{2}}}, \\ 
		& \text{RMSE log}=\sqrt{\frac{1}{\left| N \right|}\sum\limits_{i\in N}{{{\left\| \log \left( {{d}_{i}} \right)-\log \left( {{{\tilde{d}}}_{i}} \right) \right\|}^{2}}}}, \\ 
		& \text{Abs Rel}=\frac{1}{\left| N \right|}\sum\limits_{i\in N}{\frac{\left| {{d}_{i}}-{{{\tilde{d}}}_{i}} \right|}{{{{\tilde{d}}}_{i}}}}, \\ 
		& \text{Sq Rel}=\frac{1}{\left| N \right|}\sum\limits_{i\in N}{\frac{{{\left\| {{d}_{i}}-{{{\tilde{d}}}_{i}} \right\|}^{2}}}{{{{\tilde{d}}}_{i}}}}, \\ 
		& \text{Accuracies:  }\%\text{  of }{{d}_{i}}\text{ s}\text{.t}\text{. }\max \left( \frac{{{d}_{i}}}{{{{\tilde{d}}}_{i}}},\frac{{{{\tilde{d}}}_{i}}}{{{d}_{i}}} \right)=\delta <threshold \\
	\end{aligned}	
\end{equation}

\subsection{Dataset Results}
We use different resolutions for training on the KITTI dataset and compare with the SoTA methods in recent years. The quantitative comparison results are shown in TABLE \ref{tab:kitti}. TABLE \ref{tab:kitti} is mainly divided into two categories, low resolution ($640\times192$) and high resolution ($1024\times320$), and each resolution has general and lightweight methods. In low resolution, RTS-Mono outperforms most recent methods \cite{zhu2025dual} \cite{song2023unsupervised} \cite{hou2024self} \cite{li2022self} with a very low number of parameters (3.0 M).  Compared with lightweight methods, we achieved the best performance, reducing AbsRel to 0.101, SqRel to 0.690, RMSE to 4.344, and reaching 0.897 on the ${{\delta }_{1}}<1.25$. Compared with LiteMono, a lightweight work that achieved performance SoTA before, our work improved by 5.6\% on Abs Rel, 9.8\% on Sq Rel, 4.8\% on RMSE, and 4.9\% on RMSE log. Our method is also very competitive in high resolution compared to general methods \cite{zhu2025dual} \cite{li2022self}. Compared with lightweight methods, RTS-Mono is even more advanced in all aspects, reducing SqRel to 0.667 and RMSE to 4.225 at a very low cost, 6.1\% and 1.9\% higher than LiteMono, respectively. At this time, the number of parameters of LiteMono has reached 8.7 M. At the same time, that of our method is 3.0 M. These show that RTS-Mono has achieved highly competitive performance at a very low cost for high-resolution and low-resolution image data.
\begin{table*}[!t]
	\caption{Quantitative comparison of the KITTI dataset: M means training with monocular sequence, M+Se means training with monocular sequence and semantic information, and D means training with ground truth. \textbf{Bold} indicates the best metric, and the \underline{underline} indicates the second-best metric.}
	\label{tab:kitti}
	\centering
	\setlength{\tabcolsep}{2.0mm}{
		\begin{tabular}{c c c c c c c c c c c c}
			\specialrule{0em}{3pt}{3pt}
			\toprule[1pt]
			
			\multirow{2}{*}[-1pt]{Method}  &  \multirow{2}{*}[-1pt]{Year} &  \multirow{2}{*}[-1pt]{Data}&  \multirow{2}{*}[-1pt]{\#Param}&\multirow{2}{*}[-1pt]{Resolution} &\multicolumn{4}{c}{Error metric $\downarrow $}& \multicolumn{3}{c}{Accuracy metric $\uparrow $}\\
			
			\cline{6-12}
			\specialrule{0em}{1pt}{2pt}
			~&~&~ & ~&~&Abs Rel & Sq Rel&RMSE&RMSE log&$\delta <1.25$&$\delta <{{1.25}^{2}}$&$\delta <{{1.25}^{3}}$\\
			\specialrule{0.05em}{1pt}{3pt}
			Monodepth2\cite{godard2019digging}&2019&M& 32.5 M&$640\times192$             &0.110&0.831&4.642&0.187&0.883&0.962&0.982\\
			SC-SfMLearner\cite{bian2019unsupervised}   &2019&M& 34.6 M&$813\times114$     &0.114&0.813&4.642&0.191&0.873&0.960&0.982\\
			SGDepth\cite{klingner2020self}  &2020&M+Se& 16.3 M &$640\times192$    &0.113&0.835&4.693&0.191&0.979&0.961&0.981\\
			Johnston et al.\cite{johnston2020self}  &2020&M&14.3 M &$640\times192$   &0.110&0.941&4.817&0.189&0.885&0.961&0.981\\			
			CAD-Depth\cite{yan2021channel}  &2021  &M&58.3 M&$640\times192$ &0.110&0.812&4.686&0.187&0.882&0.962&0.983\\
			HR-Depth\cite{lyu2021hr}  &2021  &M&14.7 M &$640\times192$  &0.109&0.792&4.632&0.185&0.884&0.962&0.983\\
			Ling et al.\cite{ling2021unsupervised}  &2022  &M&- &$640\times192$ &0.121&0.971&5.206&0.214&0.843&0.944&0.975\\
			RA-Depth\cite{he2022ra}  &2022  &M&\underline{10.0 M}  &$640\times192$ &\textbf{0.096}&\textbf{0.632}&\textbf{4.216}&\underline{0.171}&\underline{0.903}&\textbf{0.968}&\underline{0.985}\\
			HRANet\cite{wang20223d}  &2023  &M&14.3 M   &$640\times192$ &0.109&0.790&4.656&0.185&0.882&0.962&0.983\\
			MonoFormer\cite{bae2023deep}  &2023  &M&-   &$640\times192$ &0.106&0.839&4.622&0.183&0.889&0.962&0.983\\
			HQDec\cite{wang2023hqdec}  &2023  &M&29.3 M   &$640\times192$ &\textbf{0.096}&\underline{0.654}&4.281&\textbf{0.169}&0.896&0.965&\underline{0.985}\\
			Li et al.\cite{li2022self}  &2023  &S&- &$640\times192$ &0.106&0.767&4.602&0.183&0.876&0.960&0.984\\
			DynamoDepth\cite{sun2024dynamo}  &2024  &M&- &$640\times192$ &0.112&0.758&4.505&0.183&0.873&0.959&0.984\\
			
			Hou et al.\cite{hou2024self}  &2024  &M&34.7 M   &$640\times192$ &\underline{0.098}&0.726&4.375&0.177&0.891&\underline{0.967}&\textbf{0.986}\\
			ADPDepth \cite{song2023unsupervised}  &2024  &M&114 M   &$640\times192$ &0.119&0.886&4.831&0.196&0.966&0.955&0.980\\
			TinyDepth\cite{cheng2024tinydepth}  &2024  &M&\textbf{6.2 M}   &$640\times192$ &\textbf{0.096}&0.665&\underline{4.249}&\underline{0.171}&\textbf{0.904}&\textbf{0.968}&\underline{0.985}\\
			DAG-Net \cite{zhu2025dual}  &2025  &M&16.6 M   &$640\times192$ &0.104&0.770&4.568&0.184&0.887&0.961&0.983\\
			\specialrule{0.05em}{1pt}{3pt}
			
			Lite-HR-Depth\cite{lyu2021hr}  &2021  &M&3.1 M  &$640\times192$ &0.116&0.845&4.841&0.190&0.866&0.957&0.982\\
			R-MSFM\cite{zhou2021r}  &2021  &M&3.8 M   &$640\times192$ &0.112&0.806&4.704&0.191&0.878&0.960&0.981\\
			GuideDepth\cite{rudolph2022lightweight}  &2022  &D&5.8 M   &$640\times192$ &0.142&-&5.194&-&0.799&0.941&0.982\\
			METER\cite{papa2023meter}  &2023  &M&3.3 M   &$640\times192$ &0.126&-&4.603&-&0.829&-&-\\
			
			LiteMono\cite{zhang2023lite}  &2023  &M&3.1 M  &$640\times192$ &\underline{0.107}&0.765&4.561&0.183&0.886&0.963&0.983\\
			RT-Monodepth\cite{feng2024real}  &2024  &M&2.8 M &$640\times192$ &0.125&0.959&4.985&0.202&0.857&0.952&0.979\\
			RTIA-Mono\cite{zhao2025rtia}  &2025  &M&\textbf{1.4 M} &$640\times192$ &0.118&0.858&4.810&0.196&0.865&0.955&0.980\\
			\specialrule{0.05em}{1pt}{3pt}
			RTS-Mono-XS(Ours)  &2025&M&\underline{2.4 M}    &$640\times192$   &\textbf{0.101}&\textbf{0.690}&\underline{4.357}&\underline{0.175}&\textbf{0.897}&\textbf{0.966}&\textbf{0.984}\\
			RTS-Mono-S(Ours)  &2025&M&3.0 M    &$640\times192$   &\textbf{0.101}&\underline{0.698}&\textbf{4.344}&\textbf{0.174}&\textbf{0.897}&\textbf{0.966}&\textbf{0.984}\\
			\specialrule{0.05em}{1pt}{3pt}
			Monodepth2\cite{godard2019digging}&2019&M&\underline{14.3 M}&$1024\times320$             &0.115&0.882&4.701&0.190&0.879&0.961&0.982\\
			HR-Depth\cite{lyu2021hr}  &2021  &M&14.7 M &$1024\times320$ &0.106&0.755&4.472&0.181&0.892&0.966&\underline{0.984}\\		
			CAD-Depth\cite{yan2021channel}  &2021  &M&58.3 M&$1024\times320$ &0.102&0.834&4.407&0.178&0.898&0.966&\underline{0.984}\\
			DIFFNet \cite{zhou2021self}&2021  &M&\textbf{10.9 M}&$1024\times320$  &\underline{0.097}&0.722&4.345&0.174&\underline{0.907}&\underline{0.967}&\underline{0.984}\\
			MonoViT\cite{zhao2022monovit}  &2022  &M&37 M&$1024\times320$  &\textbf{0.096}&\underline{0.714}&\underline{4.292}&\textbf{0.172}&\textbf{0.908}&\textbf{0.968}&\underline{0.984}\\
			Li et al.\cite{li2022self}  &2023  &S&- &$1024\times320$  &0.098&\textbf{0.691}&\textbf{4.283}&\underline{0.173}&0.887&0.966&\textbf{0.985}\\
			DAG-Net \cite{zhu2025dual}  &2025  &M&16.6 M   &$1024\times320$ &0.099&0.715&4.336&0.178&0.896&0.965&0.983\\
			\specialrule{0.05em}{1pt}{3pt}
			Lite-HR-Depth\cite{lyu2021hr}&2021&M&3.1 M&$1024\times320$             &0.111&0.799&4.612&0.184&0.878&0.963&\underline{0.984}\\
			R-MSFM3\cite{zhou2021r}  &2021  &M&3.5 M &$1024\times320$ &0.112&0.773&4.581&0.189&0.879&0.960&0.982\\		
			R-MSFM6\cite{zhou2021r}  &2021  &M&3.8 M&$1024\times320$ &0.108&0.748&4.470&0.185&0.889&0.963&0.982\\
			LiteMono\cite{zhang2023lite}  &2023  &M&3.1 M&$1024\times320$  &\underline{0.102}&0.746&4.444&0.179&0.896&0.965&0.983\\
			LiteMono-8M\cite{zhang2023lite}  &2023  &M&8.7 M&$1024\times320$  &\textbf{0.097}&0.710&4.309&\underline{0.174}&\textbf{0.905}&\underline{0.967}&\underline{0.984}\\
			\specialrule{0.05em}{1pt}{3pt}
			RTS-Mono-XS(Ours)  &2025&M&\textbf{2.4 M}    &$1024\times320$  &\textbf{0.097}&\textbf{0.659}&\underline{4.246}&\textbf{0.171}&\underline{0.904}&0.966&\underline{0.984}\\
			RTS-Mono-S(Ours)  &2025&M&\underline{3.0 M}   &$1024\times320$  &\textbf{0.097}&\underline{0.667}&\textbf{4.225}&\textbf{0.171}&\underline{0.904}&\textbf{0.968}&\textbf{0.985}\\			
			
			\bottomrule[1pt]
	\end{tabular}}
\end{table*}
\begin{figure*}[!t]
	\centering
	\includegraphics[width=6.8in]{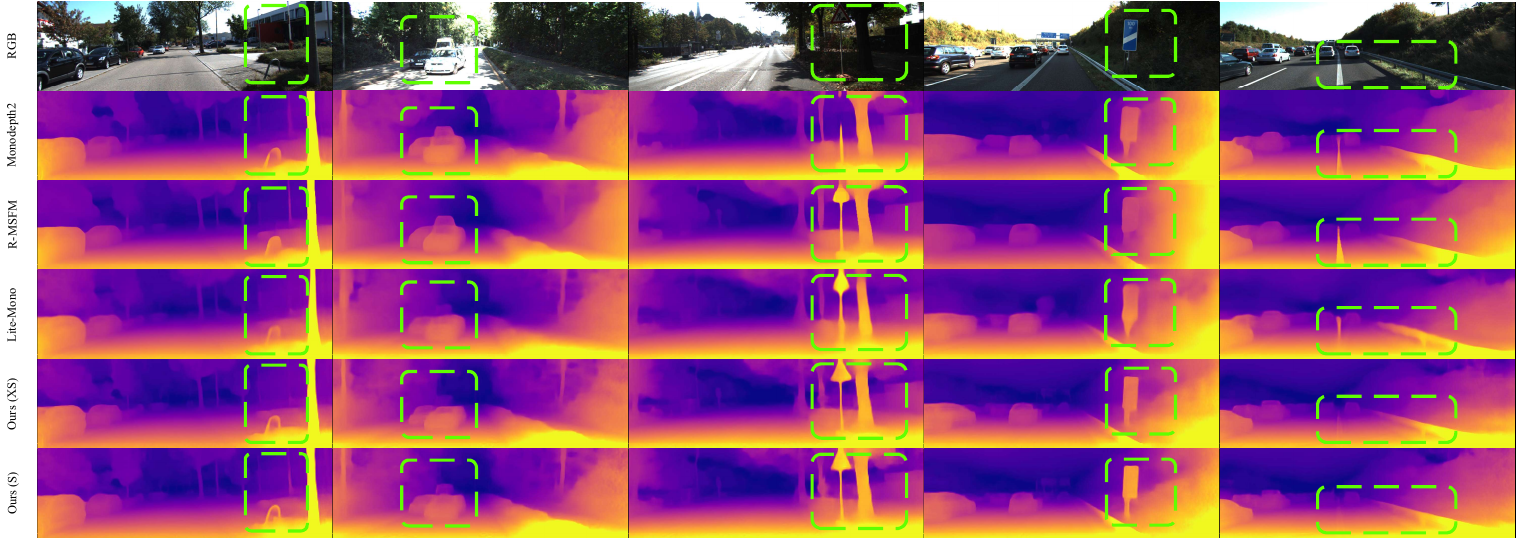}
	\caption{Qualitative comparison on the KITTI dataset.}
	\label{kitti}
\end{figure*}

In addition, the qualitative comparison of RTS-Mono with lightweight work and classic work on KITTI is shown in Fig. \ref{kitti}, in which we use green dotted boxes to mark the areas where our method has obvious advantages. It is not difficult to see that our method is more accurate in perceiving objects such as traffic signs and railings, with more restored details and sharper edges, and more accurate in perceiving the distance of objects. In the third column, the traffic sign \cite{godard2019digging} has a loss of perception, while our method can accurately perceive this sign. In the fifth column, other methods have misperceived road markings, while our method does not have such errors.

\subsection{Real-world Deployment Details}
After conducting quantitative and qualitative comparisons of the datasets, in order to verify the performance of our method in real-world deployment, we deployed RTS-Mono on an intelligent UAV system to conduct real-world depth perception experiments to verify its real-world deployment performance. Specifically, we deployed the model trained on the KITTI dataset on NVIDIA Jetson Orin, and installed NVIDIA Jetson Orin on the UAV, and combined it with a series of equipment such as a monocular camera, battery, antenna, and ground monitoring station to form an intelligent UAV depth perception system. The performance of our method is tested by real-time perception in the real world through the intelligent UAV depth perception system.

\subsection{Real-world Deployment Results}
The parameters, inference speed, accuracy, and other indicators of the network are directly related to the network's deployment capability on devices with limited computing resources. Therefore, before the deployment, we conducted comparative experiments on mobile devices, focusing on these key metrics. All methods were tested on NVIDIA Jetson Orin. We also used 697 test images from the KITTI dataset for verification and unified the resolution to $640\times192$.
 The results are shown in TABLE \ref{tab:deployment}. The inference speed of our method reached 49 FPS and was highly accurate. For \cite{he2022ra} \cite{cheng2024tinydepth}, although they are better than our method in terms of accuracy and RMSE metrics, they have slower inference speeds and more parameters. Especially in inference speed, both methods did not reach 30 FPS, and they cannot be considered to have the ability of real-time inference, which is unacceptable for mobile devices. Although \cite{rudolph2022lightweight} and \cite{feng2024real} are ahead of our method in inference speed and latency, they are far behind our method in accuracy and RMSE. Compared with the methods of the same level as ours, Lite-Mono \cite{zhang2023lite} and RTIA-Mono \cite{zhao2025rtia}, we outperform it in terms of inference speed, latency, accuracy, and RMSE.
 
\begin{table}[!t]
	\caption{Quantitative comparison of real-world deployment capability metrics. \textbf{Bold} indicates the best metric, and the \underline{underline} indicates the second-best metric.}
	\label{tab:deployment}
	\centering
	\setlength{\tabcolsep}{0.60mm}{
		\begin{tabular}{c c c c c c c }
			\specialrule{0em}{3pt}{3pt}
			\toprule[1pt]
			
			Method&Data & \#Param$\downarrow $&FPS $\downarrow $&latency $\downarrow $&$\delta <1.25$ $\uparrow $&RMSE $\downarrow $\\
			\specialrule{0.05em}{1pt}{3pt}
			RADepth   &M&10.0 M&12&83.3 ms&\underline{0.903}&\underline{4.052}\\
			MonoViT-Tiny &M&10 M&\underline{14}&71.3 ms&0.895&4.459\\
			TinyDepth&M&\underline{6.2 M}&\textbf{21}&\textbf{47.0} ms&\textbf{0.904}&\textbf{4.043}\\
			R-MSFM&M&\textbf{3.8 M}&\textbf{21}&\underline{47.6 ms}&0.878&0.704\\
			\specialrule{0.05em}{1pt}{1pt}
			GuideDepth &M&5.8 M&\underline{51}&\underline{19.6 ms}&0.799&5.194\\
			Lite-Mono &M&3.1 M&45&27.0 ms&\underline{0.886}&4.561\\
			RTIA-Mono &M&\textbf{1.4 M}&49&20.8 ms&0.865&4.810\\
			RT-MonoDepth &M&2.8 M&\textbf{83}&\textbf{12.0 ms}&0.857&4.985\\
			
			\specialrule{0.05em}{1pt}{3pt}
			RTS-Mono-XS(Ours)  &M&\underline{2.4 M}&49&20.4 ms&\textbf{0.897}&\underline{4.357}\\
			RTS-Mono-S(Ours)  &M&3.0 M&45&22.0 ms&\textbf{0.897}&\textbf{4.344}\\
			\bottomrule[1pt]
	\end{tabular}}
\end{table}

\begin{figure}[!t]
	\centering
	\includegraphics[width=3.2in]{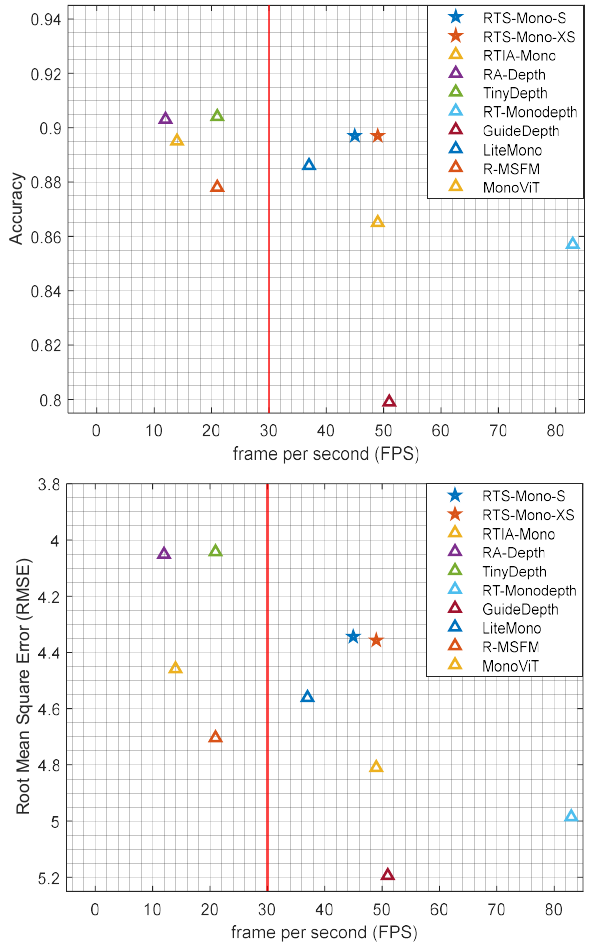}
	\caption{Accuracy vs. FPS and RMSE vs. FPS on NVIDIA Jetson Orin.}
	\label{vs}
\end{figure}

In order to more intuitively reflect whether the balance between model performance and computational efficiency is achieved, we use Accuracy vs. FPS and RMSE vs. FPS to express it, as shown in Fig \ref{vs}. The red vertical line in the figure represents the basic frame rate that achieves real-time performance. It can be seen that compared with the two methods \cite{he2022ra} \cite{cheng2024tinydepth}, although their performance is slightly better than ours, they do not meet the real-time standard and cannot perform real-time inference on mobile devices. Among the methods that are higher than the real-time running standard, our method performs much better than other methods, and the inference speed is close to 50 FPS, which is very suitable for deployment on mobile devices for real-time high-precision inference. Therefore, we believe our method has balanced performance and computational efficiency and is suitable for real-world deployment.
\begin{figure}[!t]
	\centering
	\includegraphics[width=3.5in]{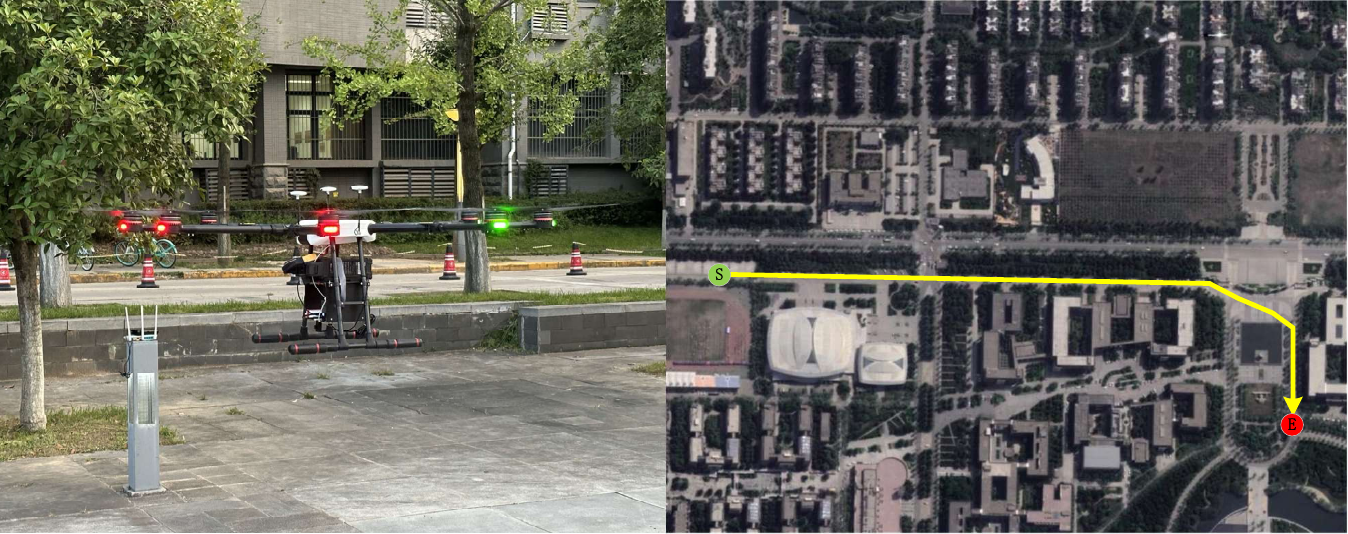}
	\caption{Real-world deployment flight process and flight route.}
	\label{realworld}
\end{figure}

\begin{figure*}[!t]
	\centering
	\includegraphics[width=6.8in]{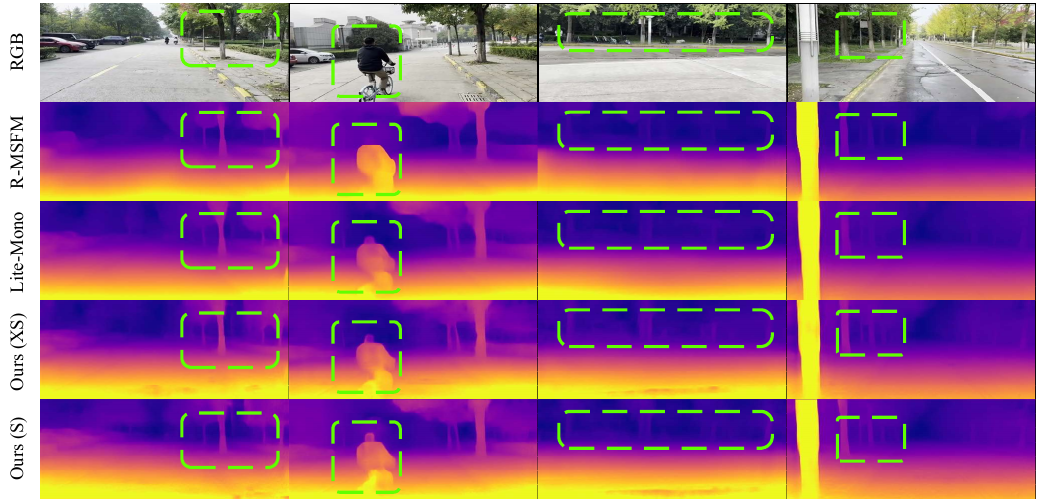}
	\caption{Qualitative comparison of real-world deployments.}
	\label{NPU}
\end{figure*}

Then, we selected a real environment, planned a route, deployed the intelligent UAV with our method, flew along the route, and performed real-time depth perception. During the flight, the resolution of the monocular camera on the UAV was set to $480\times270$. The specific flight process and flight route are shown in Fig. \ref{realworld}. It should be noted that the real environment we selected is similar to the KITTI dataset in terms of scene, weather and lighting. During the flight, not only was the depth map predicted in real-time, but the monocular visual information of the entire flight process was also recorded, and the real-time collected images and predicted depth information were saved every 100 frames. We used the recorded visual information to deploy R-MSFM and LiteMono on the UAV to predict the depth of the same scene and made a qualitative comparison with our method. The results are shown in Fig. \ref{NPU}. We use the green dotted box to mark the area where our method has obvious advantages. It can be seen that our method is more accurate in the perception of pedestrians, trees, and other objects, and there is no problem of perception loss (R-MSFM has this problem in the same scene), and it is the most accurate in controlling the overall structure of the scene and restoring the details. In summary, our method is superior to R-MSFM and LiteMono regarding real-world deployment performance.
\begin{table*}[!t]
	\caption{Ablation study of encoder-decoder architecture, \textbf{bold} indicates the best metric.}
	\label{tab:Ablation_network}
	\centering
	\setlength{\tabcolsep}{3.5mm}{
		\begin{tabular}{c c c c c c c c c}
			\specialrule{0em}{3pt}{3pt}
			\toprule[1pt]
			
			\multirow{2}{*}[-1pt]{Network}  &\multirow{2}{*}[-1pt]{\#Param}  &\multicolumn{4}{c}{Error metric $\downarrow $}& \multicolumn{3}{c}{Accuracy metric $\uparrow $}\\
			
			\cline{3-9}
			\specialrule{0em}{1pt}{2pt}
			~&~&Abs Rel & Sq Rel&RMSE&RMSE log&$\delta <1.25$&$\delta <{{1.25}^{2}}$&$\delta <{{1.25}^{3}}$\\
			\specialrule{0.05em}{1pt}{3pt}
			ResNet18 + DepthDecoder             &14.3 M &0.111&0.771&4.547&0.184&0.880&0.963&\textbf{0.984}\\
			ResNet18 + SDecoder             &\textbf{11.6 M} &\textbf{0.110}&\textbf{0.734}&\textbf{4.541}&\textbf{0.182}&\textbf{0.884}&\textbf{0.964}&\textbf{0.984}\\
			\specialrule{0.05em}{1pt}{3pt}
			ResNet50 + DepthDecoder             &32.5 M &0.104&0.716&4.430&0.178&0.891&0.965&0.984\\
			ResNet50 + SDecoder              &\textbf{28.8 M}  &\textbf{0.103}&\textbf{0.702}&\textbf{4.428}&\textbf{0.177}&\textbf{0.893}&\textbf{0.965}&\textbf{0.985}\\
			\specialrule{0.05em}{1pt}{3pt}
			
			Lite-Encoder + DepthDecoder     &3.1 M &0.103&0.700&4.398&0.177&0.890&0.965&0.984\\
			Lite-Encoder + SDecoder  (ours)         &\textbf{3.0 M}
			& \textbf{0.101}&\textbf{0.698}&\textbf{4.344}&\textbf{0.174}&\textbf{0.897}&\textbf{0.966}&\textbf{0.984}\\
			\bottomrule[1pt]
	\end{tabular}}
\end{table*}	

\subsection{Discussions}
From the above dataset and real-world deployment experiments, we can see that our proposed method has achieved competitive results with very low parameters and high efficiency. We attribute such outstanding results to two reasons. On the one hand, we designed a lightweight and efficient network architecture consisting of a lightweight encoder and an efficient decoder. The lightweight encoder ensures the overall low cost of the network, and the efficient decoder ensures that the network has real-time inference speed and reliable performance on a lightweight basis. On the other hand, we introduced a loss function different from general self-supervised learning into our training process. This loss function adds cross-scale depth consistency loss based on the conventional loss function, which significantly improves the accuracy of depth prediction. Our claims will be verified in the ablation study.

\subsection{Ablation study}
We aim to design a lightweight, efficient, real-time inference self-supervised monocular depth estimation network. Therefore, we designed a lightweight encoder-decoder architecture. The lightweight encoder lays the foundation for the entire network. The decoder chooses to use as few features and processing steps as possible to maximize the efficiency of the network and ensure performance. Our encoder-decoder architecture strives to achieve a balance between network efficiency and performance. In addition, we introduced a cross-scale consistency loss in the self-supervised training process to further improve the accuracy of depth prediction. Next, we will verify the effectiveness of each part. It should be noted that all the results of the ablation experiment were trained on the KITTI dataset at a resolution of $640\times192$. Unless otherwise specified, all results were trained using the self-supervised loss function of this paper.

\begin{table*}[!t]
	\caption{Comparison of network architecture balance. \textbf{Bold} indicates the best metric, and \underline{underline} indicates the second-best metric.}
	\label{tab:Economic}
	\centering
	\setlength{\tabcolsep}{4.5mm}{
		\begin{tabular}{c c c c c c c c c  }
			\specialrule{0em}{3pt}{3pt}
			\toprule[1pt]
			
			\multirow{2}{*}[-1pt]{Method}  & \multicolumn{2}{c}{Encoder} &\multicolumn{2}{c}{Decoder}&\multicolumn{2}{c}{Full Model}&  \multirow{2}{*}[-1pt]{FPS $\uparrow $ }&\multirow{2}{*}[-1pt]{Abs Rel $\downarrow $}\\
			
			\cline{2-7}
			\specialrule{0em}{1pt}{2pt}
			~& \#Param $\downarrow $&FLOPs $\downarrow $ & \#Param $\downarrow $&FLOPs $\downarrow $&\#Param $\downarrow $&FLOPs $\downarrow $&~&~\\
			\specialrule{0.05em}{1pt}{3pt}
			R-MSFM             &\textbf{0.7 M}  &\textbf{2.4 G}&3.1 M&28.8 G&3.8M&31.2 G&21&0.112\\
			Lite-Mono             &2.8 M  &4.3 G&\textbf{0.2 M}&\textbf{0.7 G}&3.1 M&\underline{5.1 G}&\underline{45}&\underline{0.107}\\
			RTIA-Mono             &\underline{1.2 M}  &\underline{3.0 G}&\textbf{0.2 M}&\textbf{0.7 G}&\textbf{1.4 M}&\textbf{3.7 G}&\textbf{49}&0.118\\
			Ours-XS             &2.2 M &4.0 G&\textbf{0.2 M}&\underline{2.0 G}&\underline{2.4 M}&6.0 G&\textbf{49}&\textbf{0.101}\\
			Ours-S             &2.8 M  &4.3 G&\textbf{0.2 M}&\underline{2.0 G}&3.0 M&6.3 G&\underline{45}&\textbf{0.101}\\
			\bottomrule[1pt]
	\end{tabular}}
\end{table*}

\begin{table*}[!t]
	\caption{Ablation study of self-supervised loss function. Best metrics are in \textbf{boldface}.}
	\label{tab:loss}
	\centering
	\setlength{\tabcolsep}{3.0mm}{
		\begin{tabular}{c c c c c c c c c c}
			\specialrule{0em}{3pt}{3pt}
			\toprule[1pt]
			
			\multirow{2}{*}[-1pt]{Loss Function}  &\multirow{2}{*}[-1pt]{Network}  &     \multirow{2}{*}[-1pt]{\#Param} &\multicolumn{4}{c}{Error metric $\downarrow $}& \multicolumn{3}{c}{Accuracy metric $\uparrow $}\\
			
			\cline{4-10}
			\specialrule{0em}{1pt}{2pt}
			~&~ & ~&Abs Rel & Sq Rel&RMSE&RMSE log&$\delta <1.25$&$\delta <{{1.25}^{2}}$&$\delta <{{1.25}^{3}}$\\
			\specialrule{0.05em}{1pt}{3pt}
			\multirow{6}{*}[-1pt]{Baseline}  &Monodepth2(18)& 14.3 M             &0.115&0.903&4.863&0.193&0.877&0.959&0.981\\
			~&Monodepth2(50)& 32.5 M             &0.110&0.831&4.642&0.187&0.883&0.962&0.982\\
			\cline{2-10}
			\specialrule{0em}{1pt}{2pt}
			
			~&Lite-Mono-small& 2.5 M             &0.110&0.802&4.671&0.186&0.879&0.961&\textbf{0.982}\\
			~&Ours-XS&\textbf{ 2.4 M  }           &\textbf{0.105}&0.780&\textbf{4.548}&\textbf{0.182}&\textbf{0.890}&\textbf{0.963}&\textbf{0.983}\\
			\cline{2-10}
			\specialrule{0em}{1pt}{2pt}
			
			~&Lite-Mono& 3.1 M             &0.107&0.765&4.561&0.183&0.886&0.963&\textbf{0.983}\\
			~&Ours-S& \textbf{3.0 M}             &\textbf{0.104}&0.754&\textbf{4.564}&\textbf{0.182}&\textbf{0.888}&\textbf{0.963}&\textbf{0.983}\\
			\specialrule{0.05em}{1pt}{3pt}
			\multirow{6}{*}[-1pt]{\makecell[c]{Cross-Scale \\Depth \\Consistency \\Loss}} 
			&Monodepth2(18)& 14.3 M             &0.111&0.771&4.547&0.184&0.880&0.963&0.984\\
			~&Monodepth2(50)& 32.5 M             &0.104&0.716&4.430&0.178&0.891&0.965&0.984\\
			\cline{2-10}
			\specialrule{0em}{1pt}{2pt}
			
			~&Lite-Mono-small& 2.5 M             &0.109&0.758&4.517&0.182&0.881&0.962&\textbf{0.984}\\
			~&Ours-XS& \textbf{2.4 M}             &\textbf{0.101}&\textbf{0.690}&\textbf{4.357}&\textbf{0.175}&\textbf{0.897}&\textbf{0.966}&\textbf{0.984}\\
			\cline{2-10}
			\specialrule{0em}{1pt}{2pt}
			
			~&Lite-Mono& 3.1 M             &0.103&0.700&4.398&0.177&0.890&0.965&\textbf{0.984}\\
			~&Ours-S& \textbf{3.0 M }            &\textbf{0.101}&\textbf{0.698}&\textbf{4.344}&\textbf{0.174}&\textbf{0.897}&\textbf{0.966}&\textbf{0.984}\\
			\bottomrule[1pt]
	\end{tabular}}
\end{table*}

\subsubsection{Encoder-decoder Architecture}
In order to verify the effectiveness of the encoder-decoder architecture, we designed a series of experiments to verify the effectiveness of the encoder and decoder simultaneously. The results are shown in TABLE \ref{tab:Ablation_network}. ResNet-18 and ResNet-50 represent ResNet network architectures with depths of 18 and 50, respectively. LiteEncoder is the encoder we introduced, DepthDecoder is the decoder of the classic work \cite{godard2019digging}, and SDecoder is the decoder we designed. TABLE \ref{tab:Ablation_network} is divided into three parts: the results of the encoder-decoder architecture training composed of encoders ResNet18, ResNet50, LiteEncoder, and decoders DepthDecoder and SDecoder. TABLE \ref{tab:Ablation_network} shows that with a fixed decoder, the decoder we introduced can achieve the best results in each group. With a fixed encoder, the decoder we designed can also achieve the best results, and the number of parameters of the network architecture can be significantly reduced. This series of comparative experiments can prove the effectiveness of the encoder-decoder architecture of our network.

\subsubsection{Discussion on the Balance of Network Architecture}
In order to analyze the balance of our network architecture in detail and verify whether our method has achieved a balance between performance and efficiency, we compared the network architecture and efficiency in detail with two lightweight works, \cite{zhou2021r} and \cite{zhang2023lite}, with similar sizes, as shown in TABLE \ref{tab:Economic}. As seen from TABLE \ref{tab:Economic}, the number of parameters and the amount of calculation of the R-MSFM encoder are the lowest, which means that the encoder of R-MSFM has a high computational efficiency. This is because it only uses part of the ResNet18 network architecture, which gives computational efficiency a qualitative leap, but it also inevitably leads to decreased network performance. In order to compensate for this decrease, the decoder of R-MSFM becomes very complex, and the calculation amount is enormous, which seriously affects its inference speed. The encoder of Lite Mono is consistent with ours, using a CNN-Transformer hybrid architecture, which is lightweight and efficient while also ensuring performance. Lite-Mono is very lightweight because the encoder cannot output the complete five features like ResNet18, so it uses part of the decoder of \cite{godard2019digging} for depth prediction, which cannot well guarantee local and global feature modeling. Therefore, although the efficiency of Lite-Mono is guaranteed, the network's performance still has room for improvement, as shown in the results of TABLE \ref{tab:Economic}. RTIA-Mono \cite{zhao2025rtia} has an efficient encoder and adopts a fusion upsampling process similar to our method. However, it does not have the shallow feature fusion step and simplifies the fusion upsampling process. As a result, although RTIA-Mono's computational efficiency and network size are better than our method, its network performance is far inferior to ours, which is consistent with our discussion in Section \ref{sec:discuss}.

Our method uses the same structure as Lite-Mono on the encoding side, which can ensure performance while keeping the network lightweight. On the decoding side, unlike the disadvantages of \cite{zhou2021r} and \cite{zhang2023lite} on the decoding side, we guarantee network performance and efficiency to the greatest extent with extremely low parameters and low computational complexity. The inference speed and network performance are the best compared to \cite{zhou2021r} and \cite{zhang2023lite}. Therefore, our method considers both performance and inference speed and achieves a balance between performance and computational efficiency.

\subsubsection{Loss Function}
Another reason our method achieves excellent performance is that we introduce a loss function different from the general self-supervised monocular depth estimation loss function. This loss function adds cross-scale consistency loss as an additional supervision signal based on the basic supervision signal. Therefore, we need to verify the loss function's effectiveness and our network architecture's overall effectiveness. To this end, we designed two sets of comparative experiments. In the first set, we selected a general self-supervised loss function \cite{godard2019digging} named baseline to compare the performance of each network under the baseline. In the second set, we used the loss function we introduced to train each method and compared the performance of each network under this loss function. At the same time, combining the two sets of experiments can also verify the effectiveness of the loss function we introduced. The results are shown in TABLE \ref{tab:loss}. Each set of experiments shows that our method can maintain its lead compared to the advanced method LiteMono of the same size, whether it is trained under the baseline or the loss function we introduced, which verifies the effectiveness of our network architecture under different loss functions. From the two sets of experiments, the loss function we introduced can significantly improve the network compared to the baseline, further verifying the effectiveness of the loss function we introduced.

\section{Conclusions}
This paper proposes a self-supervised monocular depth estimation method, RTS-Mono, that can be deployed in practice and perform real-time accurate inference. RTS-Mono is a lightweight and efficient encoder-decoder architecture. We introduce a lightweight CNN-Transformer hybrid architecture encoder and design a multi-scale sparse fusion decoder to maximize the inference speed and ensure network performance. We conducted dataset experiments and real-world deployment experiments. In the dataset experiments, RTS-Mono achieved SoTA performance compared with similar size methods in low resolution ($640\times192$) and high resolution ($1024\times320$) with extremely low parameter count (3 M). Specifically, we reduced Abs Rel to 0.101, RMSE to 4.344, and a1 to 0.897 in low resolution. We reduced Abs Rel to 0.097, Sq Rel to 0.667, RMSE to 4.225, and a1 to 0.904 in high resolution. In addition, we built an intelligent UAV depth perception system to test the real-world deployment performance of our method. Experiments show that RTS-Mono can perform real-time inference in the real world with an inference speed of 49 FPS, and its real-world real-time depth perception performance also surpasses state-of-the-art work of similar size. The ablation study shows the effectiveness of the network architecture we proposed. In summary, our method balances network performance and computational efficiency, laying the foundation for the practical application of depth estimation on vehicles, robots, and UAVs.

\bibliographystyle{IEEEtran}
\bibliography{IEEEabrv}

\end{document}